\documentclass[letterpaper, 10 pt, conference]{ieeeconf}  % Comment this line out if you need a4paper

\IEEEoverridecommandlockouts                              % This command is only needed if 
\usepackage{amsmath} % assumes amsmath package installed
\usepackage{amssymb}  % assumes amsmath package installed

\usepackage{graphicx}
\usepackage[tight,footnotesize]{subfigure}

\usepackage{caption}
\usepackage{subfig}
\usepackage{amsmath} % assumes amsmath package installed
\usepackage{amssymb}  % assumes amsmath package installed
\usepackage{hyphenat}
\usepackage[table]{xcolor}
\usepackage{multirow}
\usepackage{url}
\usepackage[binary-units=true]{siunitx}
\usepackage{float}
\usepackage{booktabs,amsfonts,dcolumn}
\usepackage{tabularx}

\usepackage{tikz}

\newcommand\acomment[1]{\textcolor{red}{A:#1}}
\newcommand\mcomment[1]{\textcolor{olive}{mariosx:#1}}

\newcommand\bcomment[1]{\textcolor{red}{B:#1}}

\long\def\invis#1{}

\newcommand\sect[1]{Section~\ref{#1}}
\newcommand\fig[1]{Figure~\ref{#1}}
\newcommand\tab[1]{Table~\ref{#1}}

\newcommand\etal{\textit{et al.\ }}
\newcommand\eg{\textit{e.g.,\ }}
\newcommand\ie{\textit{i.e.,\ }}

\newcommand\copyrighttext{%
  \footnotesize \textcopyright 2019 IEEE. Personal use of this material is permitted.
  Permission from IEEE must be obtained for all other uses, in any current or future
  media, including reprinting/republishing this material for advertising or promotional
  purposes, creating new collective works, for resale or redistribution to servers or
  lists, or reuse of any copyrighted component of this work in other works.
  DOI: 10.1109/IROS40897.2019}
\newcommand\copyrightnotice{%
\begin{tikzpicture}[remember picture,overlay]
\node[anchor=south,yshift=10pt] at (current page.south) {\fbox{\parbox{\dimexpr\textwidth-\fboxsep-\fboxrule\relax}{\copyrighttext}}};
\end{tikzpicture}%
}

\title{\LARGE \bf Experimental Comparison of Open Source Visual-Inertial-Based State Estimation Algorithms in the Underwater Domain }

\author{Bharat Joshi$^\dag$, Sharmin Rahman$^\dag$, Michail Kalaitzakis$^\ddag$, Brennan Cain$^\dag$, James Johnson,$^\dag$ Marios Xanthidis$^\dag$,\\ 
Nare Karapetyan$^\dag$, Alan Hernandez$^{\dag\dag}$, Alberto Quattrini Li$^{\ddag\ddag}$, Nikolaos Vitzilaios$^\ddag$, and Ioannis Rekleitis$^\dag$% <-this % stops a space
\thanks{$^\dag$Computer Science and Engineering Department, University of South Carolina, Columbia, SC, USA, {\tt\small \{bjoshi,srahman,bscain,jvj1\}@email.sc.edu, \{mariosx,nare\}@email.sc.edu, yiannisr@cse.sc.edu}}%
\thanks{$^\ddag$Mechanical Engineering Department, University of South Carolina, Columbia, SC, USA, {\tt\small michailk@email.sc.edu, vitzilaios@sc.edu}}%
\thanks{$^{\dag\dag}$Computer Science Department, MiraCosta College, Oceanside, CA USA, {\tt\small ahernandez@miracosta.edu}}%
\thanks{$^{\ddag\ddag}$Department of Computer Science, Dartmouth College, Hanover, NH, USA, {\tt\small alberto.quattrini.li@dartmouth.edu}}%
\thanks{The authors would like to thank the National Science Foundation for its support (NSF 1513203, 1637876). The authors would also like to acknowledge the help of the Woodville Karst Plain Project (WKPP) in data collection.}
}

\begin{document}

\maketitle
\copyrightnotice
\thispagestyle{empty}
\pagestyle{empty}

%%%%%%%%%%%%%%%%%%%%%%%%%%%%%%%%%%%%%%%%%%%%%%%%%%%%%%%%%%%%%%%%%%%%%%%%%%%%%%%%
\begin{abstract}
A plethora of state estimation techniques have appeared in the last decade using visual data, and more recently with added inertial data. Datasets typically used for evaluation include indoor and urban environments, where supporting videos have shown impressive performance. However, such techniques have not been fully evaluated in challenging conditions, such as the marine domain. In this paper, we compare ten recent open\hyp source packages to provide insights on their performance and guidelines on addressing current challenges. Specifically, we selected direct and indirect methods that fuse camera and Inertial Measurement Unit (IMU) data together. Experiments are conducted by testing all packages on datasets collected over the years with underwater robots in our laboratory. All the datasets are made available online.
\end{abstract}
\invis{cross validating each package on the accompanying datasets of every other package. Second, we tested them}

%%%%%%%%%%%%%%%%%%%%%%%%%%%%%%%%%%%%%%%%%%%%%%%%%%%%%%%%%%%%%%%%%%%%%%%%%%%%%%%%
\section{INTRODUCTION}
An important component of any autonomous robotic system is estimating the robot's pose and the location of the surrounding obstacles -- a process termed Simultaneous Localization and Mapping (SLAM). During the last decade the hardware has dramatically improved, both in performance, as well as cost reduction. As a result, camera sensors and Inertial Measurement Units (IMU) are now mounted in most robotic systems, and in addition to all smart\hyp devices (phones and tablets). With the proliferation of visual inertial devices many researchers have designed novel state estimation algorithms for \emph{Visual Odometry} (VO)~\cite{6096039,fraundorfer2012visual} or \emph{visual SLAM}~\cite{Fuentes-Pacheco:2015:VSL:2717330.2717465}.\invis{; see, \eg \cite{6096039,fraundorfer2012visual} for a comprehensive overview on VO, as well as a recent survey on visual SLAM methods\cite{Fuentes-Pacheco:2015:VSL:2717330.2717465}.} In this paper we will examine the performance of several state\hyp of\hyp the\hyp art Visual Inertial State Estimation open\hyp source packages in the underwater domain. 
%
%Fuentes-Pacheco \etal \cite{Fuentes-Pacheco:2015:VSL:2717330.2717465} recently surveyed Visual SLAM methods.

\begin{figure}[ht]
\begin{center}
 \includegraphics[width=0.95\columnwidth]{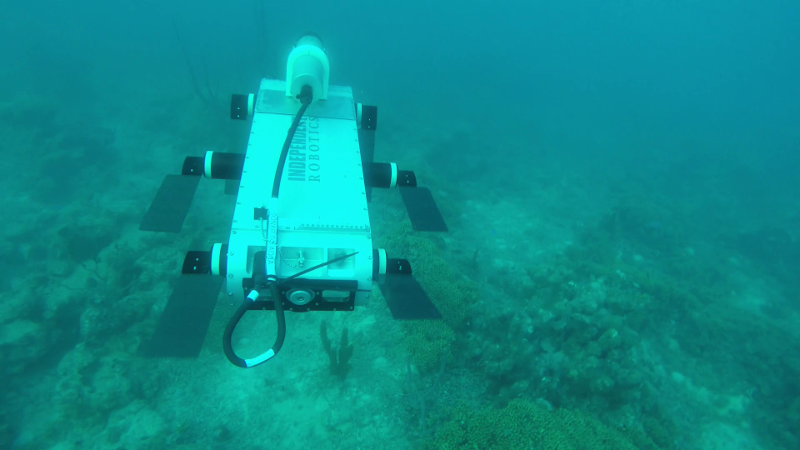}
\end{center}
\vspace{-0.1in} \caption{Data collection by an Aqua2 underwater robot~\cite{Rekleitis2005d}.\vspace{-0.2in}\invis{using a hand\hyp held sonar, stereo, inertial, and depth sensor suite}\label{fig:Beauty}}
\end{figure}

 %Here a very brief overview of the different methods for visual state estimation.
Vision based state estimation algorithms can be classified into a few broad classes.
Some methods -- such as~\cite{Davison2007,Geiger2011IV,4538852,orb-slam2} -- are based on feature detection and matching to solve the Structure from Motion (SfM) problem. The poses are estimated solving a minimization problem on the re-projection error that derives from reconstructing the tracked features. Such methods are often termed \emph{indirect methods}.
\emph{Direct methods} -- such as~\cite{lsd,dso} -- instead, use pixel intensities directly to minimize the photometric error, skipping the feature detection and matching step.

To improve the state estimation performance, data from other sensors can be fused together with the visual information. Typically, in literature there are two general approaches. One approach is based on \emph{filtering}, where IMU measurements are used to propagate the state, while visual features are considered in the update phase~\cite{mourikis2007multi,jones2011visual,Kelly01012011}. The second approach is based on \emph{tightly\hyp coupled nonlinear optimization}: all sensor states are jointly optimized, minimizing the IMU and the reprojection error terms~\cite{okvis,mur2017visual}.

Recently, there has been great interest in comparing different approaches, \eg \cite{williams2009comparison,QuattriniLiIser2016VO,chahinesurvey,Delmerico:254865}. However, typically the analysis is limited to only a few packages at a time~\cite{williams2009comparison,Delmerico:254865}, and rarely to a specific domain~\cite{chahinesurvey}. In addition, all these comparisons consider higher quality visual data, not applicable to the underwater domain. In previous work~\cite{QuattriniLiIser2016VO}, some of the authors compared a good number of state\hyp of\hyp the\hyp art open\hyp source vision\hyp based state estimation packages on several different datasets -- including ones collected by marine robots, as the one shown in \fig{fig:Beauty}. The results showed that most packages exhibited higher errors in datasets from the marine domain. 

\invis{An important consideration in our work was avoiding the inherent bias in the selection of the testing datasets. When the authors of a specific approach/package select testing datasets, they naturally exclude scenarios where the robot behavior is incompatible with the proposed approach. For example, a drone based approach often requires a motion in all three axes to initialize, however, a ground vehicle cannot move in the Z and Y axis. In this work, we have selected a wide variety of scenarios, considering a diverse set of vehicles used to acquire the testing datasets, and using a mixture of publicly available datasets and datasets collected by our lab. }

This paper considerably expands the previous analysis~\cite{QuattriniLiIser2016VO} by selecting ten recent open source packages, focused towards the integration of visual and inertial data, and testing them on new datasets, made publicly available\footnote{\label{note1}\url{https://afrl.cse.sc.edu/afrl/resources/datasets/}}. The contribution of this paper is not in providing a new state estimation method, but in taking a snapshot of the current capabilities of the most recent open\hyp source packages, by evaluating them on \invis{simulated underwater benchmarks~\cite{duarte2016towards} and primarily in} datasets collected over the years with underwater robots and sensors deployed by the Autonomous Field Robotics Lab. Such a snapshot allows us to draw some insights on weaknesses, strengths, and failure points of the different methods. A general guideline can be distilled to drive the design of new robust state estimation algorithms.
 
\invis{This paper is structured as follows. The next section gives a brief overview of the tested algorithms. \sect{sec:data} illustrates the datasets used for the comparison. \sect{sec:res} shows the results and \sect{sec:disc} provides related insights. Finally, \sect{sec:conclusion} concludes the paper.}

\section{Related Work and Methods Evaluated}

Visual state estimation methods can be classified according to the following criteria:
\begin{itemize}
 \item \emph{number of cameras}, e.g., monocular, stereo, or more rarely multiple cameras; 
 \item \emph{the presence of an IMU}, where its use could be \emph{optional} for some methods or \emph{mandatory} for others;
 \item \emph{direct} vs.\ \emph{indirect} methods;
\item \emph{loosely} vs.\ \emph{tightly\hyp coupled} optimization when multiple sensors are used -- e.g., camera and IMU;
 \item the presence or lack of a \emph{loop closing} mechanism.
\end{itemize}
In the following, we provide a brief description of the algorithms evaluated in this paper; please refer to the original papers for in depth coverage. 
While these algorithms do not cover the exhaustive list of algorithms in the literature, they cover approaches, along the dimensions mentioned above. \invis{consisting of loosely and tightly coupled approaches, direct and feature based methods and filtering and optimization based algorithms.}

\invis{\paragraph{DPPTAM}
Dense Piecewise Planar Tracking and Mapping (DPPTAM)~\cite{dpptam} is a direct visual odometry algorithm for dense reconstruction using a monocular camera. Dense reconstruction is based on detecting planar regions -- assumed to be homogeneous-color regions. The TUM RGB-D Dataset~\cite{sturm12iros} was used to assess the performance of the proposed approach in the original paper.}

\paragraph{LSD-SLAM}
Direct Monocular SLAM is a direct method that operates on intensities of images from a monocular camera~\cite{lsd} both for tracking and mapping, allowing dense 3D reconstruction. Validated on custom datasets from TUM, covering indoor and outdoor environments. \invis{While working in data from above water deployments this package consistently diverged in underwater data, as was also reported earlier~\cite{QuattriniLiIser2016VO}}

\paragraph{DSO}
Direct Sparse Odometry~\cite{dso} is a new direct method proposed after LSD-SLAM by the same group. It probabilistically samples pixels with high gradients to determine the optical flow of the image. DSO minimizes the photometric error over a sliding window. Extensive testing with the TUM monoVO dataset~\cite{tum-mono-dataset} validated the method.

\paragraph{SVO 2.0}

Semi-Direct Visual Odometry~\cite{svo} relies on both a direct method for tracking and triangulating pixels with high image gradients and a feature-based method for jointly optimizing structure and motion. It uses the IMU prior for image alignment and can be generalized to multi-camera systems. The proposed system has been tested in a lab setting with different sensors and robots, as well as the EuRoC~\cite{Burri25012016} and ICL-NUIM~\cite{handa:etal:ICRA2014} datasets.

\paragraph{ORB-SLAM2}
ORB-SLAM2~\cite{orb-slam2} is a monocular/stereo SLAM system, that uses ORB features for tracking, mapping, relocalizing, and loop closing. It was tested in different datasets, including KITTI \cite{Geiger2013IJRR} and EuRoC \cite{Burri25012016}. The authors extended it to utilize the IMU~\cite{mur2017visual}, although, currently, the extended system is not available open source.

\paragraph{REBiVO}
Realtime Edge Based Inertial Visual Odometry~\cite{rebivo} is specifically designed for Micro Aerial Vehicles (MAV). In particular, it tracks the pose of a robot by fusing data from a monocular camera and an IMU. The approach first processes the images to detect edges to track and map. An EKF is used for estimating the depth. \invis{The system was evaluated using the EuRoC dataset~\cite{Burri25012016}. The reliance to edge detection resulted in consistent failure over all underwater datasets.}

\paragraph{Monocular MSCKF}
An implementation of the original Multi-State Constraint Kalman Filter from Mourikis and Roumeliotis \cite{mourikis2007multi}\footnote{MSCKF was first utilized with underwater data in \cite{ShkurtiIROS2011} operating off\hyp line with recorded data.} was made available as open source from the GRASP lab~\cite{msckf-danilidis}. It uses a monocular camera and was tested on the EuRoC dataset~\cite{Burri25012016}.

\paragraph{Stereo-MSCKF}
Stereo Multi-State Constraint Kalman Filter~\cite{msckf-kumar-ral} is also based on MSCKF \cite{mourikis2007multi} and was made available from another group of the GRASP lab, while using a stereo camera, and has comparable computational cost as monocular solutions with increased robustness. Experiments in the EuRoC dataset and on a custom dataset collected with a UAV show good performance.

\paragraph{ROVIO}
Robust Visual Inertial Odometry~\cite{rovio} employs an Iterated Extended Kalman Filter to tightly fuse IMU data with images from one or multiple cameras. The photometric error is derived from image patches that are used as landmark descriptors and is included as residual for the update step. The EuRoC dataset~\cite{Burri25012016} was used for assessing the performance of the system.

\paragraph{OKVIS}
Open Keyframe-based Visual-Inertial SLAM~\cite{okvis} is a tightly-coupled nonlinear optimization method that fuses IMU data and images from one or more cameras. Keyframes are selected according to spacing rather than considering time-successive poses. The optimization is performed over a sliding window and states out of that window are marginalized. Experiments with a custom-made sensor suite validated the proposed approach.

\paragraph{VINS-Mono}
VINS-Mono \cite{Lin17} estimates the state of a robot equipped with an IMU and monocular camera. The method is based on a tightly-coupled optimization framework that operates with a sliding window. The system has also loop detection and relocalization mechanisms. Experiments were performed in areas close to the authors' university and the EuRoC dataset~\cite{Burri25012016}.

\tab{tab:methods} lists the methods evaluated in this paper and their properties.
\begin{table}[bh]
 \centering
 \caption{Summary of characteristics for evaluated methods.\label{tab:methods}}
 \resizebox{\columnwidth}{!}{
 \begin{tabular}{lcccccc}
  \multicolumn{2}{c}{\textbf{Method}} & \textbf{Camera} & \textbf{IMU} & \textbf{Indirect/} & \textbf{(L)oosely/} & \textbf{Loop } \\
                                      &                 &              & &\textbf{Direct} & \textbf{(T)ightly} & \textbf{Closure} \\
  \invis{DPPTAM~\cite{dpptam} & mono & no  & direct & N/A & no  \\}
  LSD-SLAM&\cite{lsd} & mono & no & direct & N/A & yes  \\
  DSO&\cite{dso} & mono & no & direct & N/A & no \\
  SVO&\cite{svo} & multi & optional & semi\hyp direct & N/A & no \\
  ORB-SLAM2&\cite{orb-slam2} & mono, stereo & no  & indirect & N/A & yes  \\
  REBiVO&\cite{rebivo} & mono & optional & indirect & L & no \\
  Mono-MSCKF&\cite{msckf-danilidis} & mono & yes & indirect & T & no \\
  Stereo-MSCKF&\cite{msckf-kumar-ral} & stereo & yes & indirect & T & no \\
  ROVIO&\cite{rovio} & multi & yes  & direct & T & no \\
  OKVIS&\cite{okvis} & multi & yes & indirect & T & no \\
  VINS-Mono&\cite{vins-mono} & mono & yes & indirect & T & yes \\
 \end{tabular}
 }\vspace{-0.3in}
\end{table}

\section{DATASETS}
\label{sec:data}
% \subfigure[]{\includegraphics[trim=0.0in 0.0in 0.0in 0.4in,clip,height=1.45in] {figures/bus1_photo}\label{fig:EnvA}}
%\begin{figure*}[h]
%\begin{center}
%\leavevmode
%\begin{tabular}{c}

\begin{figure*}[ht]
\begin{center}
\leavevmode
\begin{tabular}{ccc}
 \subfigure[]{\includegraphics[height=.17\textheight]{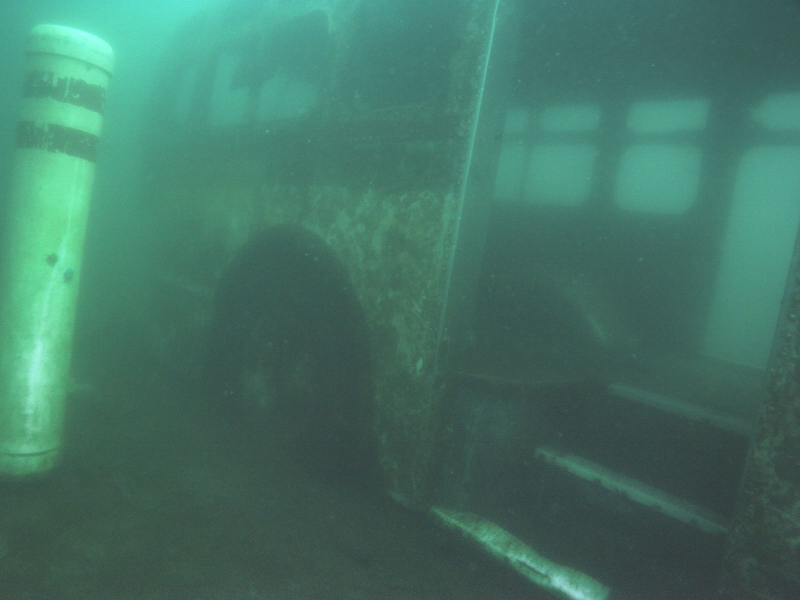}}&
 \subfigure[]{\includegraphics[height=.17\textheight]{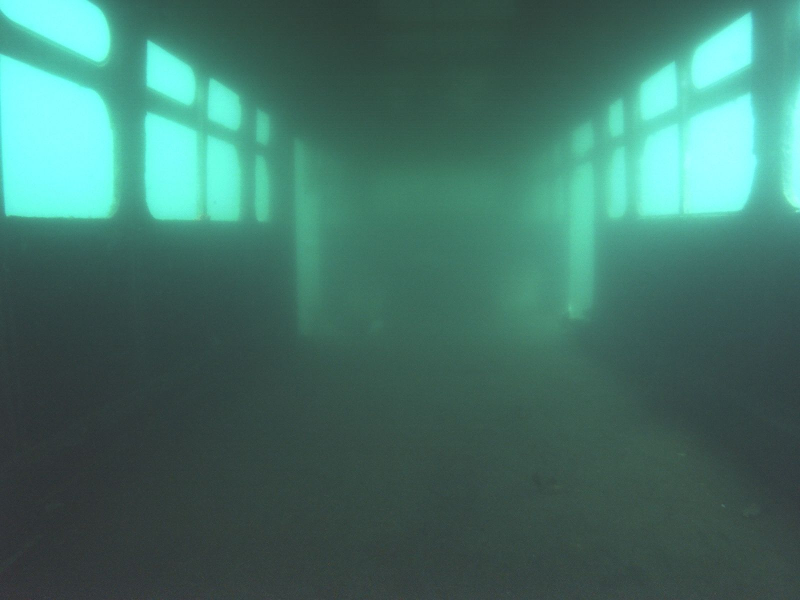}}&
 \subfigure[]{\includegraphics[height=.17\textheight]{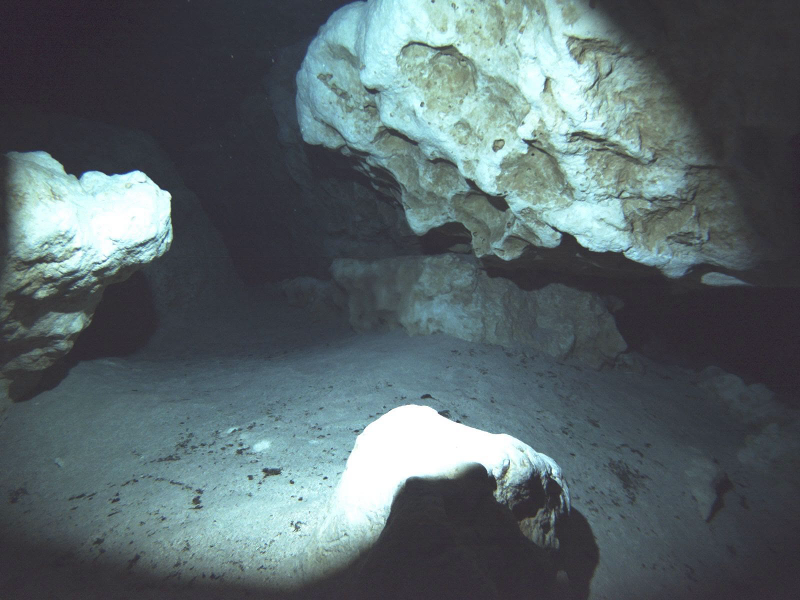}}\\
 \subfigure[]{\includegraphics[height=.17\textheight]{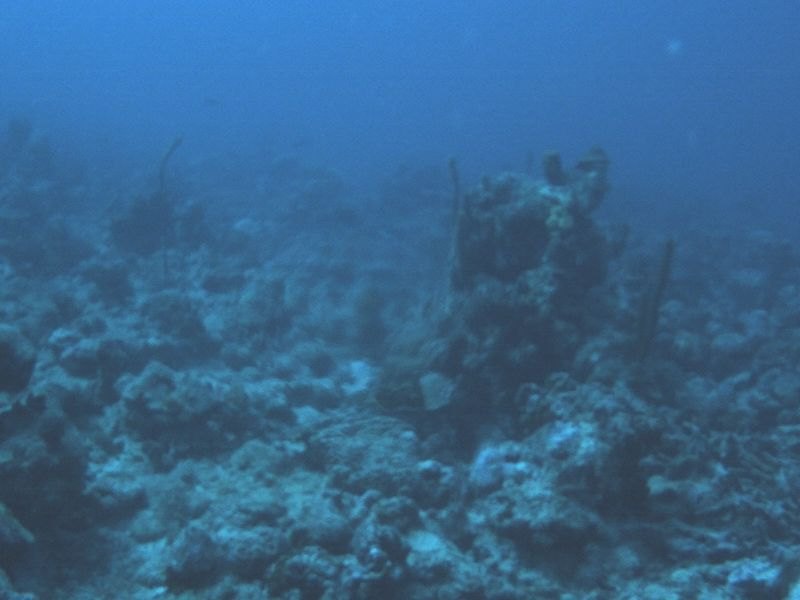}}&
 \subfigure[]{\includegraphics[height=.17\textheight]{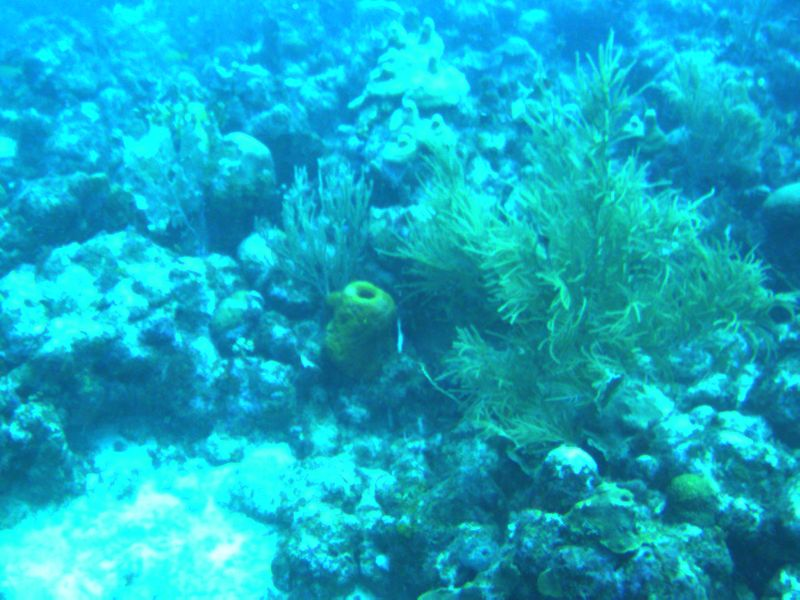}}&
 \subfigure[]{\includegraphics[height=.17\textheight]{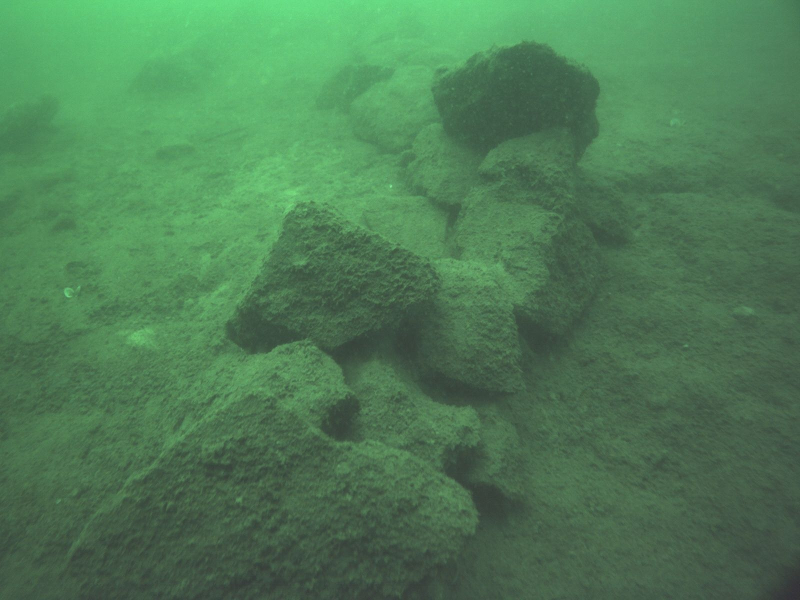}}
\end{tabular}
\end{center}
 \vspace{-0.1in}  \caption{Sample images from the evaluated datasets. (a) UW sensor suite outside a sunken bus (NC); (b) UW sensor suite inside a sunken bus (NC); (c) UW sensor suite inside a cave (FL); (d) UW sensor suite mounted on a Diver Propulsion Vehicle (DPV) over a coral reef; (e) Aqua2 AUV~\cite{Rekleitis2005d} over a coral reef; (f) AUV over a fake cemetery (SC).\vspace{-0.1in} }%\vspace{-0.3in}
  \label{fig:ds}
\end{figure*}

Most of the standard benchmark datasets represent only a single scenario, such as a lab space (\eg \cite{sturm12iros,Burri25012016}), or an urban environment (\eg Kitti~\cite{Geiger2013IJRR}), and with high visual quality. The limited nature of the public datasets is one of the primary motivations to evaluate these packages with datasets collected by our lab over the years in more challenging environments, such as underwater.

\begin{figure}[ht]
\begin{center}
 \includegraphics[width=0.8\columnwidth]{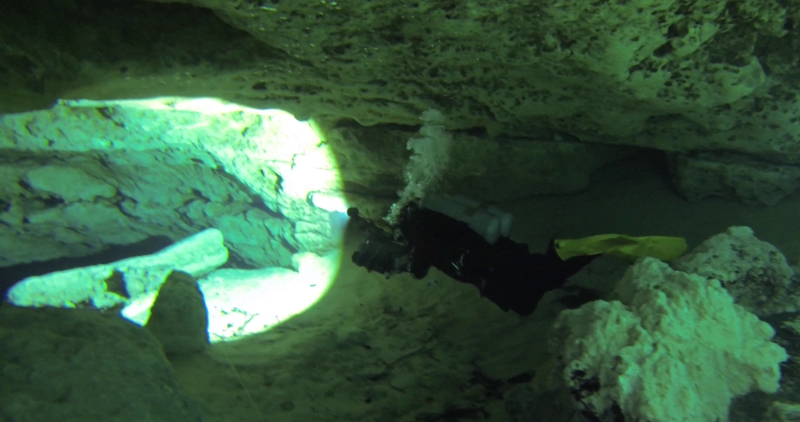}
\end{center}
 \vspace{-0.1in} \caption{Data collection using a hand\hyp held sonar, stereo, inertial, and depth sensor suite hand\hyp held inside a cavern.\vspace{-0.1in} \label{fig:rigGinnie}}
\end{figure}

\begin{figure}[ht]
\begin{center}
 \includegraphics[width=0.8\columnwidth]{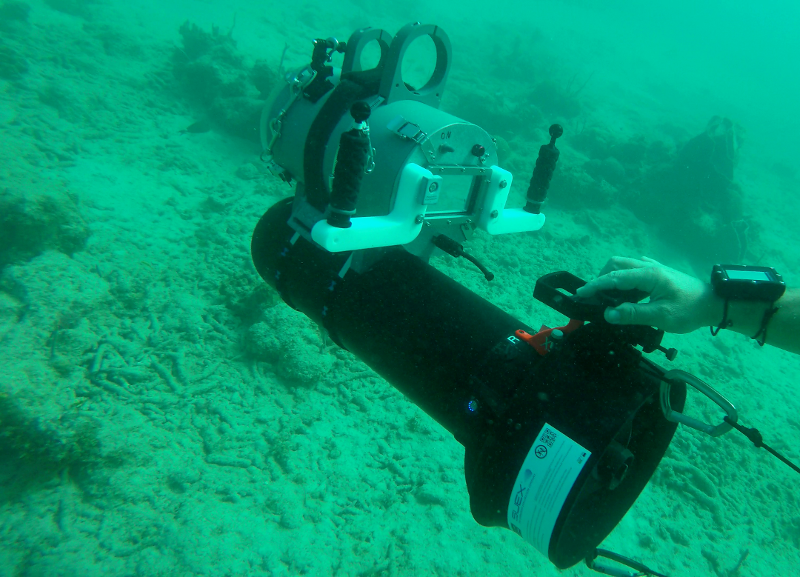}
\end{center}
\vspace{-0.1in}  \caption{Data collection using a hand\hyp held sonar, stereo, inertial, and depth sensor suite mounted on a DPV. Operations over the coral reefs of Barbados. \vspace{-0.1in} \label{fig:rigdpv}}
\end{figure}

In particular, the datasets used can be categorized according to the robotic platform used:
\begin{itemize}
\item Underwater sensor suite~\cite{RahmanOceans2018} operated by a diver around a sunken bus (Fantasy Lake, North Carolina) -- see \fig{fig:ds}(a),(b) -- and inside an underwater cave (Ginnie Springs, Florida); see \fig{fig:ds}(c). The custom-made underwater sensor suite is equipped with an IMU operating at \SI{100}{\Hz} (MicroStrain 3DM-GX15) and a stereo camera running at 15 fps, $1600 \times 1200$ (IDS UI-3251LE); see \fig{fig:rigGinnie} for operations at Ginnie Springs.
\item Underwater sensor suite~\cite{RahmanOceans2018} mounted on an Diver Propulsion Vehicle (DPV); see \fig{fig:rigdpv}. Data collected over the coral reefs of Barbados; see \fig{fig:ds}(d).
\item Aqua2 Autonomous Underwater Vehicle (AUV)~\cite{Rekleitis2005d} over a coral reef (\fig{fig:ds}(e)) and an underwater structure (Lake Jocassee, South Carolina) (\fig{fig:ds}(f)), with the same setup as the underwater sensor suite. \invis{Data collected by an Aqua2 Autonomous Underwater Vehicle (AUV), with the same setup as the underwater sensor suite; see \fig{fig:ds}(f).}
\end{itemize}

Note that the datasets used are more challenging, in terms of visibility, than those used in~\cite{QuattriniLiIser2016VO}, however the IMU data was collected at a higher frequency -- \ie at least \SI{100}{\Hz}.

To facilitate portability and benchmarking, the datasets were recorded as ROS bag files\footnote{\url{http://wiki.ros.org/Bags}} and have been made publicly available\footnotemark[1]. The intrinsic and extrinsic calibration parameters for the cameras and the IMU are also provided.

\section{Results}
\label{sec:res}

\subsection{Setup}
Experiments were conducted for each dataset on a computer with an Intel i7-7700 CPU @ 3.60GHz, \SI{32}{\giga\byte} RAM, running Ubuntu 16.04 and ROS Kinetic. \invis{In addition, an Intel NUC with i5-6260U CPU @ 1.80GHz, \SI{16}{\giga\byte} RAM, with similar configuration was used to verify the performance of each package. }
For each sequence in a dataset, we manually set each package parameters, by initializing them to the package's default values and then by tuning them to improve the performance. This process followed any available suggestion from the authors. Each package was run multiple times and the best performance is reported.
% For the benchmark datasets every package was run between five to ten times and the mean error value was presented. 
\invis{In addition, we selected a long trajectory (631m) from which twenty segments of ninety seconds duration were randomly selected, and the mean error for each package of the twenty segments was calculated. Overall, the quality of a trajectory is estimated using the the pose estimate is logged for each package. }
For the datasets where ground truth is available, the resulting trajectory is aligned with the ground truth using the \textit{sim(3)} trajectory alignment according to the method from \cite{Umeyama:1991:LET:105514.105525}: the similarity transformation parameters (rotation, translation, and scaling) are computed so that the least mean square error between estimate/ground\hyp truth locations, which are temporally close, is minimized. 
Note that this alignment might require a temporal interpolation as a pose estimate might not have an exact matching ground truth pose. In particular, bilinear interpolation is used.
The resulting metric is the root mean square error (RMSE) for the translation.
\invis{
Note that, if a method fails to initialize in an experimental run, that run is not included in the results, and we restart the experiment. 
}
% If a package cannot initialize repeatedly after changing the different parameters or if it tracks for less than 85\% of the total number of frames, we mark the result for the corresponding dataset as a failure.
%When ground truth is not available -- e.g., in the underwater datasets -- a qualitative evaluation is provided. 

%In addition, we also log the CPU and memory utilization at a rate of \SI{1}{\Hz}. The CPU utilization is measured as the percentage of a single core, thus yielding a CPU usage of possibly more than 100\% given that experiments are performed on a multi-core system. The memory utilization is measured as the percentage of the total RAM available -- i.e., \SI{32}{\giga\byte}. The average of the logged CPU and memory utilization over the experimental run is reported. These metrics give insights on how well a method can run on an embedded system mounted on a robot.

%\cite{RahmanICRA2018,RahmanICRA2019}

\invis{
for a large segment of the trajectory, it is considered a failure.
 However, some packages failed to initialize even after many trials. 
The pose estimates given by such packages were recorded after initialization if only few frames (\textless 5\% of the total number of frames) are skipped. Moreover, 
}

Since the datasets used contain stereo images and IMU data, we performed the evaluation of the algorithms considering the following combinations: monocular; monocular with IMU; stereo; and stereo with IMU, based on the modes supported by each VO/VIO algorithm. This not only compares the performance of various VIO algorithms, but also provides insight on how the performance\invis{, CPU load, and memory utilization} changes as  data from different sensors are fused together. \invis{Not every package is VIO right? There are some VO only. If the VO are not used then we can remove them from the discussion, or clearly say the ones we test, otherwise it seems we test everything. No we test both!}

% All of the tested approaches use some publicly available datasets, in particular, one of them that is common to all the packages is EuRoC. We validated our setting by first testing them on EuRoC, and the results were consistent with the results reported. Only DPPTAM was tested by the authors on a smaller scale dataset, TUM RGB-D. \acomment{finish comment}

\subsection{Evaluation}
\begin{figure*}[ht]
% trim={<left> <lower> <right> <upper>}
  \begin{center}
     \subfigure{\includegraphics[height=.15\textheight]{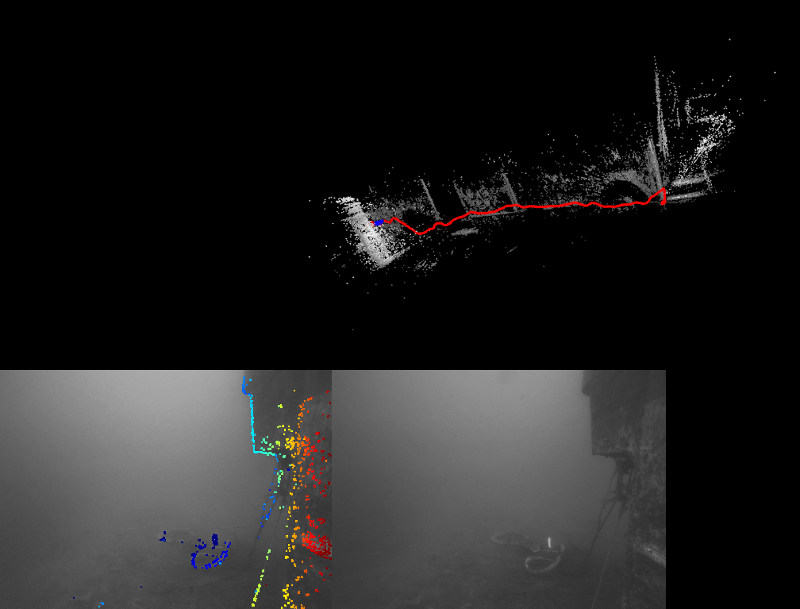}}~
     \subfigure{\includegraphics[height=.15\textheight]{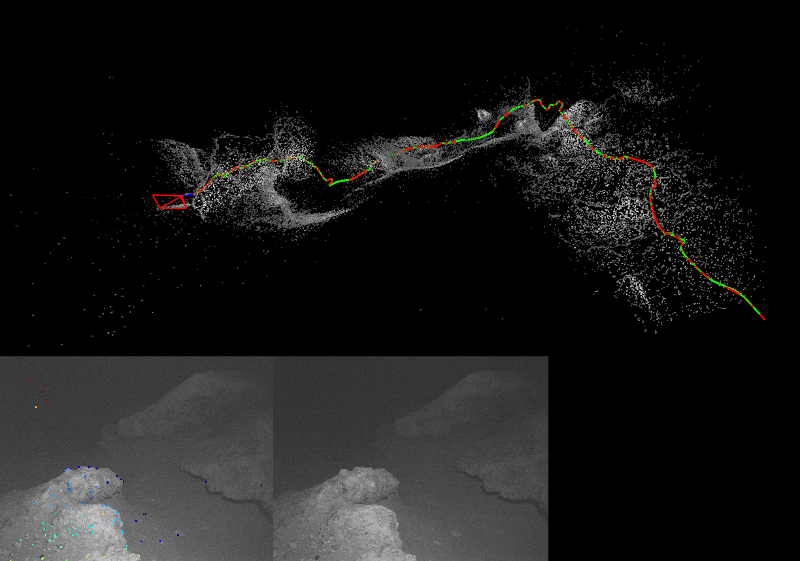}}~
     \subfigure{\includegraphics[height=.15\textheight]{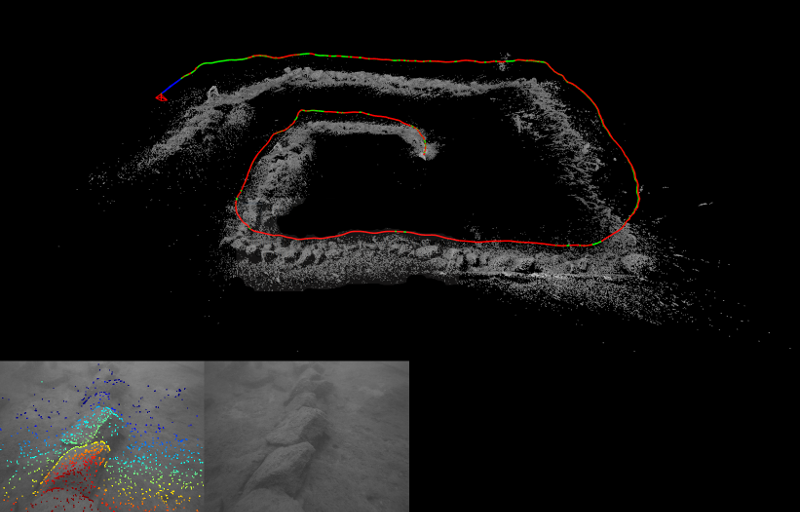}}\\
     \subfigure{\includegraphics[height=.18\textheight]{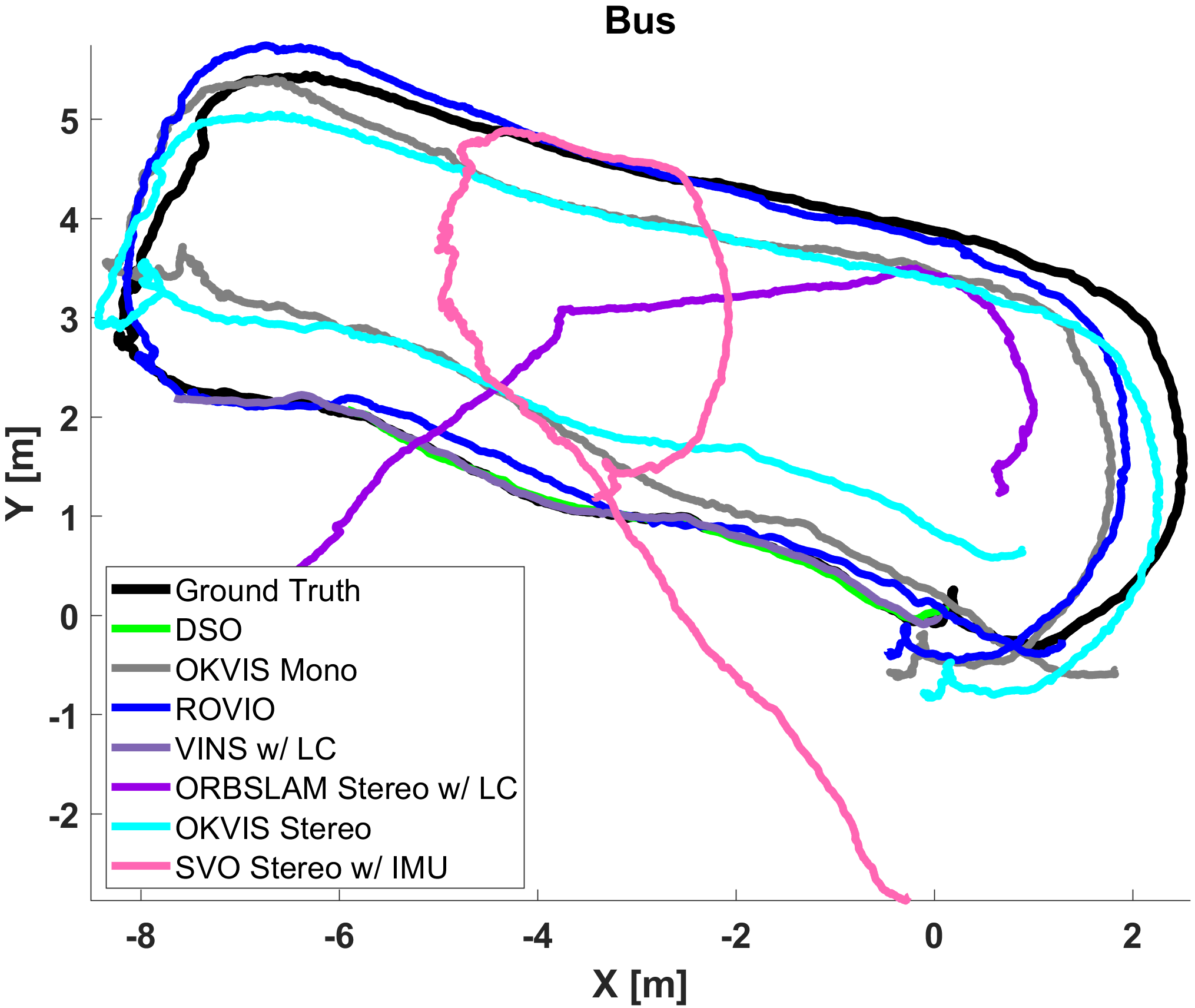}}~
     \subfigure{\includegraphics[height=.18\textheight]{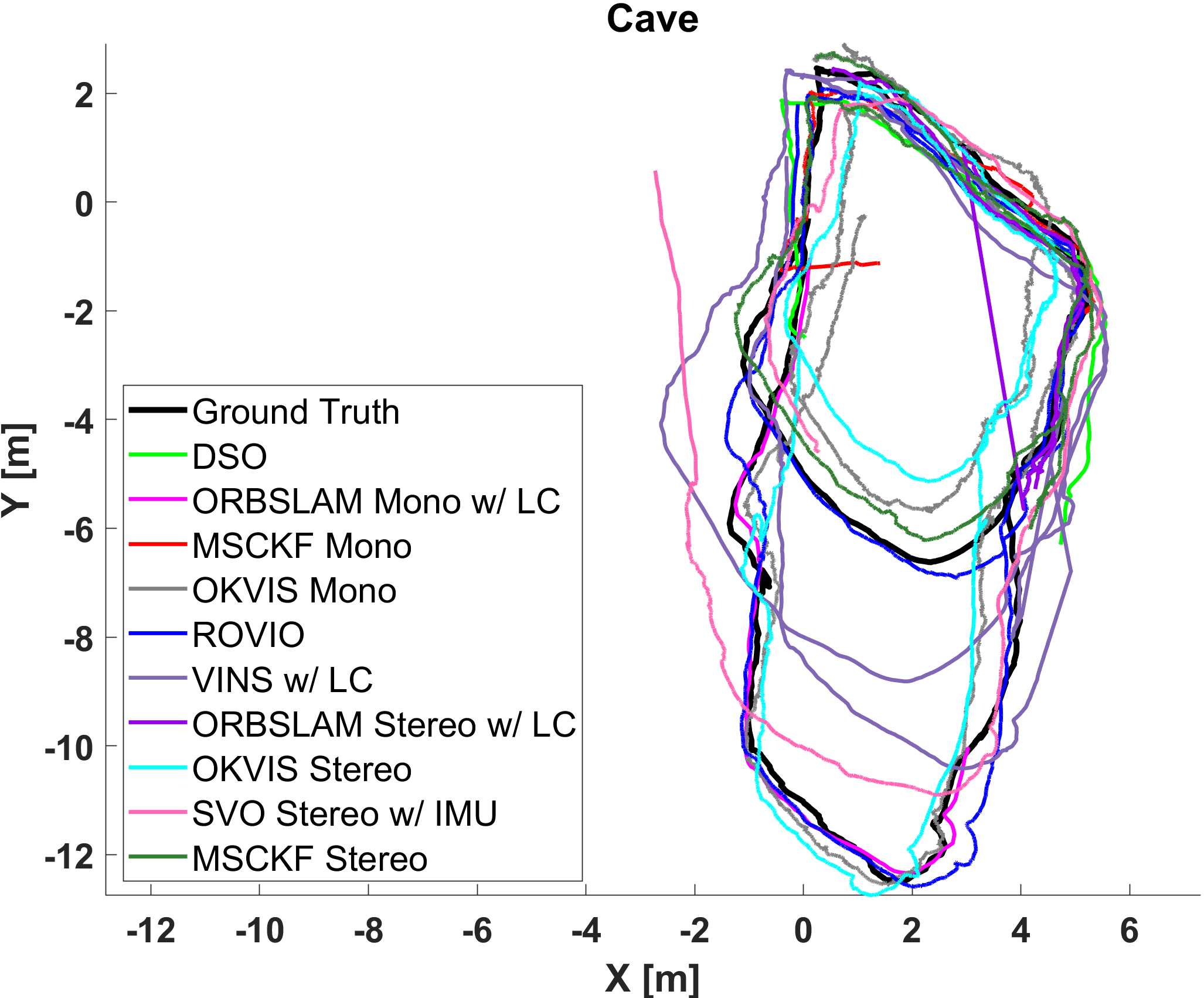}}~
     \subfigure{\includegraphics[height=.18\textheight]{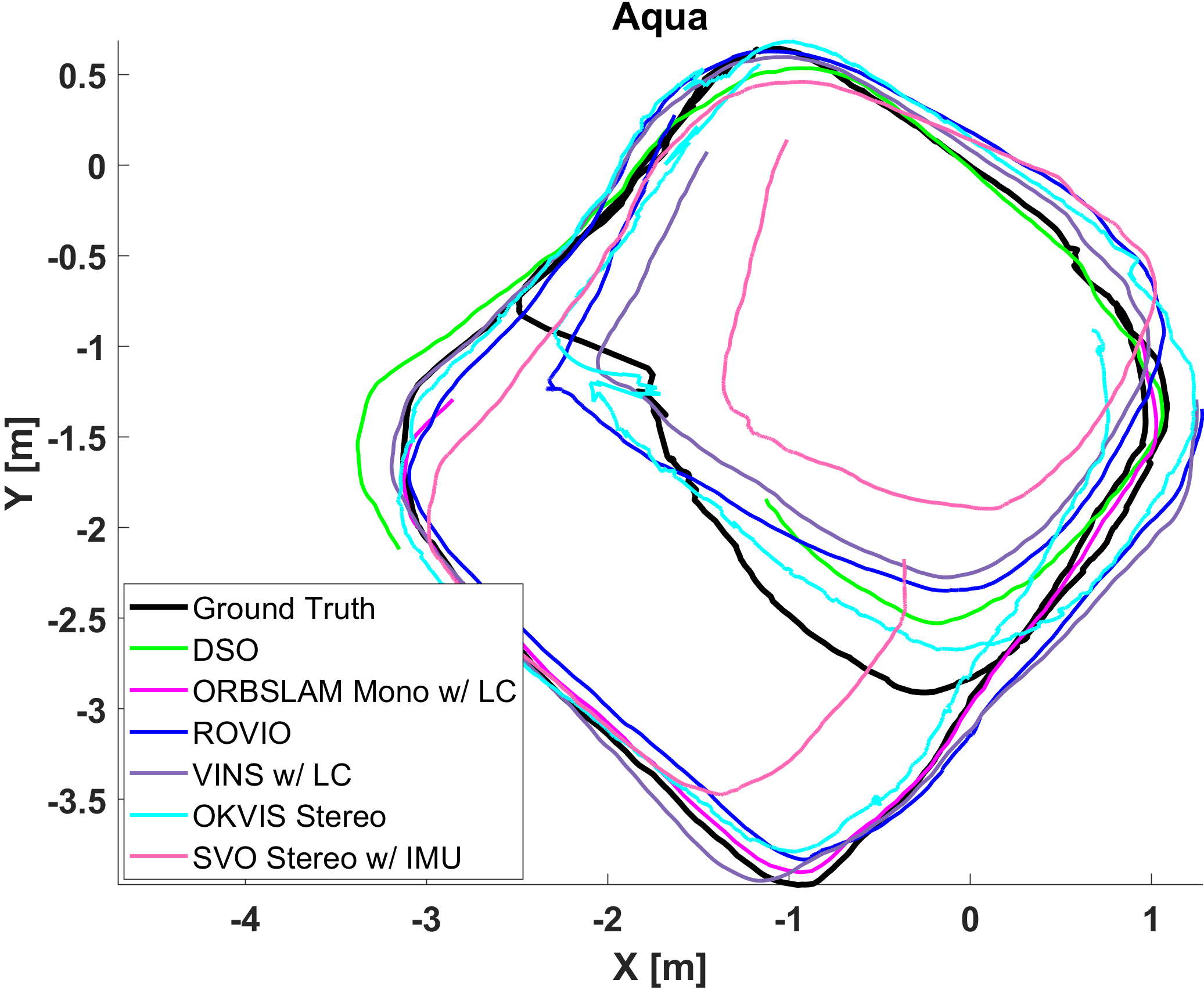}}~
     %\subfigure{\includegraphics[trim={8cm 1cm 1cm 3cm}, clip,height=.15\textheight]{figures/underwater/orb2_speedo1}\label{fig:res7}}
  \end{center}
  \vspace{-0.1in}
  \caption{Top row: The partial trajectory and reconstructions from the DSO package for three underwater datasets; outside a sunken bus; inside a cave; AUV over a fake cemetery. Bottom row: sample trajectories for the datasets of the first row.\vspace{-0.1in} }%\vspace{-0.2in}
  \label{fig:res}
\end{figure*}

% Quantitative results are reported for H/In and H/Out,  where the final trajectory obtained using gmapping\footnote{\url{http://wiki.ros.org/gmapping}} and GPS information, respectively.
For the underwater datasets, the work of Rahman \etal \cite{RahmanICRA2018,rahman2018svin2}, which fuses visual, inertial, depth and sonar range data, was used as a reference trajectory given the accurate measurements from the sonar sensor, providing an estimate of the ground truth. In addition, manual measurements between selected points have validated the accuracy of the ground truth measurements used. 

% An evaluation of the results from the underwater datasets are shown in the top rows of \tab{tab:AFRL_dataset}. The UGV datasets, indoor and outdoor, provided an easy challenge for most packages. The performance was also tested on twenty random segments of 90 sec each (average trajectory length 52m) out of the longer (631m) trajectory. It is worth noting that starting at random places degraded the performance of ORB-SLAM2. For monocular vision, ORB-SLAM2 performed better, with additional improvements over loop closure. The introduction of the IMU improved performance, and SVO was better on average; the Mono-MSCKF open source implementation failed to initialize in multiple instances. When stereo vision was used alone, ORB-SLAM2 had the best performance, with improvements when loop closure was used. Finally, stereo with IMU demonstrated improved performance by OKVIS.  

For the first four datasets (Bus/In, Bus/Out, Cave, Aqua2Lake) \invis{it should be clear this naming with the description of the datasets above. Is coral reef used as well? Let's use consistent names.} all four (acoustic, visual, inertial, and depth) sensor data were available, together with ground truth measurements. The last two datasets (DPV, Aqua2Reef) had only Visual Inertial and Depth information, and due to the complexity of the trajectory not manual measurements of ground truth. All the datasets provided a major challenge and several packages failed to track the complete trajectory. The best over several attempts was used each time, and the percentage reported indicates what part of the trajectory was completed before the package failed. 

\invis{Pure monocular vision failed in all cases.}
ORB-SLAM2 \cite{orb-slam2} using only one camera (monocular) was able to track the whole trajectory inside the cavern (Cave) and of Aqua2 over the coral reef (Aqua2Reef). Apart from that, pure monocular vision failed to track the whole trajectory in most cases. It is worth noting that DSO, while not able to track the complete trajectory, provided very detailed reconstructions of the environment where it was able to track for a period of time; see \fig{fig:ds} top. The addition of IMU brought success to OKVIS, ROVIO, and VINS-Mono (100\%), with ROVIO having the best performance, except over the coral reef where it failed due to the lack of features. OKVIS was able to track the complete trajectory in all datasets both in monocular and stereo mode. ORB-SLAM2 with loop closure in stereo mode was also able to track the whole duration on all datasets. SVO was also able to keep track in stereo mode but the resulting trajectory deviates from the ground truth. Stereo-MSCKF was able to keep track most of the time with fairly accurate trajectory. \invis{When stereo was used SVO and OKVIS completed the trajectory.} Figure \ref{fig:ds} bottom presents the trajectories of several packages together with the ground truth from~\cite{rahman2018svin2}; all trajectories were transformed to provide the best match.

The overall evaluation of all the experiments can be seen at the bottom rows of \tab{tab:AFRL_dataset}.
Specifically, we color-coded the qualitative evaluation:

\begin{itemize}
    \item{\emph{Green} means that the package was able to track at least 90\% of the path and the trajectory estimate was fairly accurate with minor deviations.}
    \item{\emph{Yellow} shows that the robot was able to localize for most of the experiment (more than 50\%). Also, the resulting trajectory might deviate from the ground truth but overall the shape of the trajectory is maintained. Yellow is also used if the robot tracked the whole sequence, but the resulted trajectory deviates significantly from the ground truth}
    \item{\emph{Orange} is used when a method tracked the robot pose for less than 50\% of the trajectory.}
    \item{\emph{Red} marks the pair of package\hyp dataset for which the robot could not initialize or diverges before tracking, or localized less than 10\% of the time. }
\end{itemize}

%\acomment{To comment on the table, showing the best, the worse. Some strengths, and weaknesses. The comments on each package should be referring directly to some datasets.}

\begin{table*}
\centering
\resizebox{\textwidth}{!}{
\begin{tabular}[t]{@{} l|ccc|cccccc|ccc|ccc}
\toprule
 &\multicolumn{3}{c|}{\textbf{Monocular}} & \multicolumn{6}{c|}{\textbf{Monocular + IMU}}  & \multicolumn{3}{c|}{\textbf{Stereo}} & \multicolumn{3}{c}{\textbf{Stereo + IMU}}\\
 & \rotatebox{90}{dso} &  \rotatebox{90}{orbslam} & \rotatebox{90}{orbslamlc} &  \rotatebox{90}{msckf} & \rotatebox{90}{okvis}  & \rotatebox{90}{rovio}  & \rotatebox{90}{svo} & \rotatebox{90}{vinsmono} & \rotatebox{90}{vinsmonolc}  & \rotatebox{90}{orbslam} & \rotatebox{90}{orbslamlc} & \rotatebox{90}{svo} &  \rotatebox{90}{okvis} & \rotatebox{90}{svo} & \rotatebox{90}{stereomsckf}\\
 
\specialrule{2pt}{2pt}{0pt}

    \multirow{2}{*}{Bus/Out (53m)} & 0.04 &  	$\times$ & 	$\times$ & 	 	$\times$ & 	1.17 & 	0.7	 & 0.19 & 	2.39 & 	2.06 & 	1.78 & 	1.38 & 	2.53 & 	0.78 & 	2.68 & 	0.61\\
    & 17\% &  $\times$ & $\times$ & $\times$ & 100\% & 100\% & 19\% & 100\% & 100\% & 88\% & 100\% & 100\% & 100\% & 100\% & 100\%\\
 \midrule
   
    \multirow{2}{*}{Cave (97m)}  & 1.56 & 0.79 & 0.78 & 2.42 & 0.92 & 0.76 & 0.79 & 3.38 & 1.48 & 0.58 & 0.59 & 3.84  & 0.98 & 1.39 & 0.66 \\
    & 29\% & 98\%	 & 100\% & 	18\% & 	100\% & 	85\% & 	31\% & 	100\% & 	100\% & 	100\% & 	100\% & 	100\% & 	100\% & 	52\% & 	100\% \\
\midrule

    \multirow{2}{*}{Aqua2Lake (55m)}  & 0.27  & 	0.64  & 	0.87  & 	$\times$  & 1.07  & 	0.45  & $\times$  & 	0.50 & 	0.50	  & 0.99  & 	0.27  & 1.61  & 	0.51  & 	1.42  & 	0.51 \\
    &23\% & 	52\% & 	57\% & 	$\times$ & 100\%  & 	100\% & $\times$ & 	100\% & 	100\% & 	69\% & 	96\% & 63\% & 	97\% & 	90\% & 	84\% \\

\specialrule{2pt}{2pt}{0pt}
   
   Bus/Out & \cellcolor{orange} & \cellcolor{red} & \cellcolor{red} & \cellcolor{red} & \cellcolor{green} & \cellcolor{green} & \cellcolor{orange} & \cellcolor{yellow} & \cellcolor{green} & \cellcolor{yellow} & \cellcolor{green} & \cellcolor{yellow} & \cellcolor{green} & \cellcolor{yellow} & \cellcolor{green} \\

   Cave & \cellcolor{orange} & \cellcolor{green} & \cellcolor{green} & \cellcolor{orange} & \cellcolor{green} & \cellcolor{yellow} & \cellcolor{orange} & \cellcolor{yellow} & \cellcolor{green} & \cellcolor{green} & \cellcolor{green} & \cellcolor{yellow} & \cellcolor{green} & \cellcolor{yellow} & \cellcolor{green} \\
   
   Aqua2Lake & \cellcolor{orange} & \cellcolor{yellow} & \cellcolor{yellow} & \cellcolor{red} & \cellcolor{green} & \cellcolor{green} & \cellcolor{red} & \cellcolor{green} & \cellcolor{green} & \cellcolor{yellow} & \cellcolor{green} & \cellcolor{orange} & \cellcolor{green} & \cellcolor{green} & \cellcolor{yellow} \\
   
   \specialrule{2pt}{2pt}{0pt}
   
    Bus/In & \cellcolor{green} & \cellcolor{red} & \cellcolor{red} & \cellcolor{red} & \cellcolor{green} & \cellcolor{green} & \cellcolor{green} & \cellcolor{yellow} & \cellcolor{yellow} & \cellcolor{green} & \cellcolor{green} & \cellcolor{green} & \cellcolor{green} & \cellcolor{green} & \cellcolor{green} \\
    DPV & \cellcolor{red} & \cellcolor{orange} & \cellcolor{orange} & \cellcolor{red} & \cellcolor{green} & \cellcolor{green} & \cellcolor{orange} & \cellcolor{yellow} & \cellcolor{yellow} & \cellcolor{yellow} & \cellcolor{yellow} & \cellcolor{orange} & \cellcolor{green} & \cellcolor{orange} & \cellcolor{orange} \\
    \invis{DPV 2& \cellcolor{cyan} & \cellcolor{orange} & \cellcolor{orange} & \cellcolor{red} & \cellcolor{cyan} & \cellcolor{cyan} & \cellcolor{orange} & \cellcolor{green} & \cellcolor{green} & \cellcolor{cyan} & \cellcolor{cyan} & \cellcolor{orange} & \cellcolor{cyan} & \cellcolor{orange} & \cellcolor{orange} \\}
    Aqua2Reef & \cellcolor{green} & \cellcolor{green} & \cellcolor{green} & \cellcolor{red} & \cellcolor{green} & \cellcolor{red} & \cellcolor{green} & \cellcolor{red} & \cellcolor{red} & \cellcolor{green} & \cellcolor{green} & \cellcolor{yellow} & \cellcolor{green} & \cellcolor{green} & \cellcolor{red} \\
\bottomrule
\end{tabular}}
\caption{Performance of the different open source packages. \invis{\acomment{specify what the first number and second number mean directly here. Specify also what x means.}} Datasets: UW sensor suite outside a sunken bus (Bus/Out); UW sensor suite inside a cave (Cave); Aqua2 (AUV) over a fake cemetery (Aqua2Lake) at Lake Jocassee; UW sensor suite inside a sunken bus (Bus/In); UW sensor suite mounted  on a Diver Propulsion Vehicle over a coral reef (DPV); Aqua2 AUV over a coral reef (Aqua2Reef). Quantitative: for each dataset, the first row specifies the RMSE after \emph{sim3} trajectory alignment; the second row is the percentage of time the trajectory was tracked; The $\times$ symbol refers to failures. Qualitative: the color chart legend is: red--failure; orange--partial failure; yellow--partial success; green-success.}~\vspace{-0.3in}
\label{tab:AFRL_dataset}
\end{table*}

\begin{figure}[ht]
\begin{center}
 \includegraphics[width=\columnwidth]{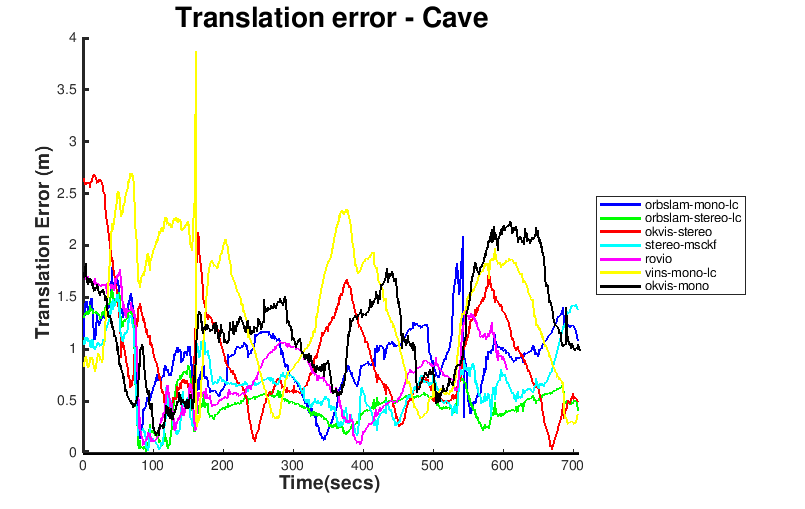}
\end{center}
\vspace{-0.1in}  \caption{Translation error  with respect to time after \textit{sim3} alignment for Cave dataset. \vspace{-0.2in} \label{fig:error_wrt_time}}
\end{figure}

%\bcomment{Alberto: can you check this paragraph?}
Figure \ref{fig:error_wrt_time} shows the absolute translation error over time in the Cave dataset. The error does not monotonously increase because of the loop closure happening at the starting point, reducing thus the accumulated error.
\invis{
The plots do not show a monotonous increase in error: The Cave dataset consists of three loops inside the cave and trajectories are aligned around the starting position at the cave entry point even after two loops. Thus, the error is less at those positions and more at positions inside the deep cave. Also, package with loop closure sometimes lose track or diverge resulting in large error resulting in large error at some positions. This error can reduce later loop closure. These factors contribute to non-monotonous nature of error with respect to time.
}

The overall performance of the tested packages is discussed next. LSD-SLAM~\cite{lsd}, REBiVO~\cite{rebivo}\invis{, Dense Piece\hyp wise Planar Tracking and Mapping (DPPTAM)~\cite{dpptam} this is the first and last mention to DPPTAM. It is not included neither on the introduction or the results}, and Monocular SVO were unable to produce any consistent results, as such, they were excluded from Table \ref{tab:AFRL_dataset}.
\invis{In the following, we comment on some of the packages that we only tested in this paper, highlighting some of their behaviors and causes for success/failure.}

%\acomment{try to merge considerations together. Do not comment for each package. Clearly do not delete, as it could be useful maybe later on, just invis it.}

%\acomment{LSDSLAM?}

%%%DSO
DSO~\cite{dso} requires full photometric calibration accounting for the exposure time, lens vignetting and non-linear gamma response function for best performance. Even without photometric calibration, it worked well on areas having high intensity gradients and when subjected to large rotations. In addition, it provided excellent reconstructions; however, in the areas with low gradient images, it was able to spatially track the motion only for a few seconds. Scale change was observed in low gradient images which can be accounted for the incorrect optimization of inverse depth. DSO requires more computational power and memory usage compared to the other packages, which is justifiable since it uses a direct method for visual odometry. \invis{As a result, it makes it unfavorable for use in mobile devices where computational power is limited.}

%LSD-SLAM~\cite{lsd} performed relatively well for the benchmark datasets, however, in more challenging environments failed to track the pose of the system.

\invis{\bcomment{Stereo DSO Section}  
  Stereo DSO \cite{wang2017stereoDSO} did not perform well on the our stereo datasets. to be completed.
}  
SVO 2.0~\cite{svo} was able to track the camera pose over long trajectories, even in parts with few features. \invis{However, while running with mono camera, it was subject to depth scale changes where tracking failed.} It tracks features using the direct method by creating a depth scene. In case of low gradient images, it was subject to depth scale changes, which was predominant in mono camera where tracking failed. SVO in stereo mode without inertial measurements was able to track most of the time but was subject to rotation errors. SVO stereo with IMU was able to keep track most of the time generating a good trajectory estimate.

%\acomment{ORBSLAM?}
ORB-SLAM2~\cite{orb-slam2} (mono) could not initialize in both datasets collected of the sunken bus (Bus/In, Bus/Out). During initialization, it calculates the homography matrix in case of a planar surface and the fundamental matrix from the 8-point algorithm for non-planar 3D structure simultaneously and chooses one of them based on inliers. Since the bus datasets are turbid with low contrast, ORB-SLAM2 cannot find any of the matrices with good certainty and both are rejected. ORB-SLAM2 works fine in the other datasets, but, without loop closure, loses track in some cases. With loop closure, despite track loss, the method can reliably relocalize after the robot traverses a previously seen region. \invis{This makes ORB-SLAM2 more robust to track loss.}\invis{Figure \ref{fig:res} the last image shows the trajectory of the AUV over the underwater structure. }

%%% REBVO comment from Brennan %%%
%REBiVO~\cite{rebivo} provided good results for the situation in which it was intended, a MAV in a busy environment. REBiVO kept track of both scaling and orientation when used on the MAV with the assistance of an IMU in an indoor environment. The main issues this package faced were with translations of slow moving robots and environments in which there were few edges to be used. Underwater and land vehicles did not benefit from this package. 

\invis{\begin{figure}[H]
	\begin{center}
      \includegraphics[height=0.2\textheight]{figures/rig1_fuzzy_edges}
      \caption{Fuzzy boundaries which make localization through Edge-Based methods difficult.}
      \label{fig:fuzzy-edges}
    \end{center}
\end{figure}
}
%%% END REBVO comment from Brennan %%%

%\acomment{MSCKF?}
Mono-MSCKF~\cite{msckf-danilidis} performed well when the AUV or sensor suite are standing still so that the IMU could properly initialize\invis{there was a stand still time for imu initialization available}, otherwise it did not track. Moreover, it was among the most efficient in terms of CPU and memory usage.%; see \fig{fig:euroc_usage}.

%%% ROVIO Comment from Mike %%%
ROVIO~\cite{rovio} is one of the most efficient packages tested. Its overall performance was robust on most datasets even when just a few good features were tracked. On the Aqua2Reef dataset though, not enough features were visible and thus it could not track the trajectory. \invis{while in cases with fewer strong features the normal operation would be disrupted. One important remark is that, when ROVIO was tested using stereo images, the monocular version would always outperform the stereo one in all aspects. Finally, the package can be easily configured, making its use in a multitude of platforms and environments possible.}
%%% END ROVIO comment from Mike %%%

%%% OKVIS comment from Sharmin
OKVIS~\cite{okvis} provided good results for both monocular and stereo.\invis{For Aqua dataset, though OKVIS monocular with IMU could not initialize and keep track while for stereo with IMU it tracked well.} In Bus/Out, despite the haze and low-contrast, OKVIS detected good features and tracked them successfully. Also in Cave, it kept track successfully and produced an accurate trajectory even in the presence of low illumination.

%%% VINS-Mono comment from Sharmin
VINS-Mono~\cite{vins-mono} works well in good illumination where there are good features to track. It was one of the few packages that worked successfully in the underwater domain. In the case of Aqua2Reef, it could not detect and track enough features and diverged. With the loop-closure module enabled, VINS-Mono reduced the drift accumulated over time in the \emph{pose} estimate and produced a globally consistent trajectory.

Stereo-MSCKF \cite{msckf-kumar} uses the Observability Constrained EKF (OC-EKF) \cite{ocekf}, which does not heavily depend on an accurate initial estimation. Also, the camera poses in the state vector can be represented with respect to the inertial frame instead of the latest IMU frame so that the uncertainty of the existing camera states in the state vector is not affected by the uncertainty of the latest IMU state during the propagation step. As a result, Stereo-MSCKF can initialize well enough even without a perfect stand still period. It uses the first 200 IMU measurements for initialization and is recommended to not have fast motion during this period. Stereo-MSCKF worked acceptably well in most datasets except Aqua2Reef and DPV. The Stereo-MSCKF could not initialize well over the coral reef due to the fast motion from the start and the low number of feature points. On the DPV dataset, it was able to track only a quarter of the full trajectory before diverging. \invis{diverged after one-fourth time was completed.}

\section{DISCUSSION}
Underwater state estimation has many open challenges, including visibility, color attenuation~\cite{SkaffBMVC2008}, floating particulates, blurriness, varying illumination, and lack of features~\cite{oliver2010image}. Indeed, in some underwater environments, there is a very low visibility that prevents seeing objects that are only a few meters away. This can be observed for example in Bus/Out, where the back of the submerged bus is not clearly visible. Such challenges make underwater localization very challenging, leaving an interesting gap to be investigated in the current state of the art. 
In addition, light attenuates with depth, with different wavelengths of the ambient light being absorbed very quickly -- e.g., the red wavelength is almost completely absorbed at \SI{5}{\m}. This alters the appearance of the image, which affects feature tracking, even in grayscale.
\invis{The appearance of color underwater is different than above, including the color loss with depth. One way some VO packages detect features is through colored regions. Color-loss, is often a challenging issue for such color-based techniques. The dominant blue intensities beyond certain depths are not allowing for color-gradient patterns to be exploited, leading to a decreased performance of such packages due to lack of features.}

The appearance of color underwater is different than above, including the color loss with depth. There is a concern when most color shifts to blue, there is a loss of sharpness, which further degrades performance. This will be a venue for further research in the future, in order to investigate the effect of any color restoration to the state estimation process.  

From the experimental results it was clear that direct VO approaches are not robust as there are often no discernible features. As such DSO and SVO, quite often fail to track the complete trajectory, however, they had the best reconstructions for the tracked parts. Similar approaches that depend on the existence of a specific feature, such as edges, are not appropriate in underwater environments in general. Overall, as expected, stereo performed better than monocular, the introduction of loop closure enabled the VO/VIO packages to track for longer periods of time, and the introduction of inertial data improved the scale estimations. 

One of the most disconcerting findings, which resulted in testing each package multiple times and reporting the best results in this paper, was the inconsistent performance under the same conditions. More specifically, most packages optimize the estimation to achieve frame rate performance, often by restricting the number iterations of RANSAC. While in theory RANSAC converges to the correct solution, in practice, with limited number of iterations allowed, the results vary widely. Most state estimation approaches, historically are tested in recorded data and the performance is presented, however, when relying on the state estimate to follow a trajectory or reach a desired pose, divergence of the estimator will have catastrophic results for the autonomous vehicle. Consequently as the best performance over a number of trials is reported, small variations are not significant.

In the results presented above, the resulting trajectory has been transformed and scaled to produce the best possible estimates. This is acceptable when ground truth is available and the process is off\hyp line, however, if a AUV has to follow a transect for a number of meters or perform a grid search, incorrect scale in the pose estimate will result in failed missions. It is worth noting here that while VINS\hyp Mono produced among the best performances, the scale was always inaccurate; a result of the monocular vision.

For increased robustness and accuracy, the OKVIS and ROVIO gave good performance, ORB\hyp SLAM with loop closure and SVO, for stereo data, and VINS\hyp Mono up to scale also performed well. DSO had the best reconstructions when it was able to track the trajectory.

\invis{Another issue is that of color filtering. One way some VO packages detect features is through colored regions. -- Color-loss, occurring in underwater environments, can be a challenging issue for such color-based techniques. The dominant blue intensities beyond certain depths are not allowing for color-gradient patterns to be exploited, leading to a decreased performance of such packages due to lack of features.} \invis{The color filtering effect of water makes these color-based techniques impossible to use and essentially reduces the RGB colorspace to a monochrome of blue. The lack of features inhibits the performance of such packages. In UW/Out, there is a short period of time in which there are insufficient color gradients, due to fine sand and water.} \invis{expand. In particular, this comparison includes the IMU. What are the improvements and still the weaknesses with VIO?}

\section{CONCLUSION}\label{sec:conclusion}
In this paper, we compared several open source visual odometry packages, with an emphasis on those that also utilize inertial data. The results confirm the main intuition that incorporating IMU measurements drastically lead to higher performance, in comparison to the pure VO packages, thus extending the results reported in~\cite{QuattriniLiIser2016VO}.\invis{ It is worth noting that sometimes the addition of a second camera did not always improve the performance. Not sure what this sentence means. } IMU improved performance was shown across a wide range of different underwater environments -- including man-made underwater structure, cavern, coral reef. 
Furthermore, the study compared popular packages using quantitative and qualitative criteria and offered insights on the performance limits of the different packages. The computational needs for online applications, were qualitative assessed.\invis{there are no quantitative results on the computational needs. Unless we provide some data, I would change the statement on the computational need to be more qualitative.}
In conclusion, for the datasets tested, OKVIS, SVO, ROVIO and VINS-Mono exhibited the best performance. 

A major concern on integrating these VIO packages in an actual robot for closed-loop control is their non\hyp determinism. \invis{the non determinism was not commented at all in the results or discussion. It should be there first.}
In pursuit of higher performance, most packages produce different output for the same input run under the same conditions. Most times the generated trajectory is reasonable, but on occasion tracking is lost and the estimator diverges or outright fails. If an autonomous vehicle was relying on the estimator to complete a mission, it would be hopelessly lost. 

Future work will include testing new packages, include new robotic platforms (\eg BlueROV2), and collecting more challenging datasets with more test cases for each package.

\bibliographystyle{IEEEtran}

\end{document}